\documentclass[letterpaper, 10 pt, conference]{ieeeconf}  % Comment this line out if you need a4paper

\IEEEoverridecommandlockouts                              % This command is only needed if you want to use the \thanks command

\usepackage{graphics} % for pdf, bitmapped graphics files
\usepackage{epsfig} % for postscript graphics files
\usepackage{mathptmx} % assumes new font selection scheme installed
\usepackage{times} % assumes new font selection scheme installed
\usepackage{amsmath} % assumes amsmath package installed
\usepackage{amssymb}  % assumes amsmath package installed
\usepackage{hyperref}
\usepackage{todonotes}

\let\labelindent\relax
\usepackage{enumitem}
\usepackage{array}
\newcolumntype{P}[1]{>{\centering\arraybackslash}p{#1}}

\usepackage{xcolor}

\newcommand{\red}[1]{\textcolor{red}{#1}}
\newcommand{\green}[1]{\textcolor{green}{#1}}
\newcommand{\blue}[1]{\textcolor{blue}{#1}}

\newcommand\copyrighttext{%
  \footnotesize \textcopyright 2024 IEEE.  Personal use of this material is permitted.  Permission from IEEE must be obtained for all other uses, in any current or future media, including reprinting/republishing this material for advertising or promotional purposes, creating new collective works, for resale or redistribution to servers or lists, or reuse of any copyrighted component of this work in other works.}
\newcommand\copyrightnotice{%
\begin{tikzpicture}[remember picture,overlay]
\node[anchor=south,yshift=10pt] at (current page.south) {\fbox{\parbox{\dimexpr\textwidth-\fboxsep-\fboxrule\relax}{\copyrighttext}}};
\end{tikzpicture}%
}

\title{\LARGE \bf
RobotGraffiti: An AR tool for semi-automated construction of workcell models to optimize robot deployment
}

\author{Krzysztof Zieli\'nski$^{1,2}$, Ryan Penning$^{2}$, Bruce Blumberg$^{2}$, Christian Schlette$^{1}$, Mikkel Baun Kjærgaard$^{1}$
\thanks{$^{1}$Authors are with
the Faculty of Engineering, the Maersk Mc Kinney Moller Institute, University of Southern Denmark
        {\tt\small krzi, chsch, mbkj@mmmi.sdu.dk }}%
\thanks{$^{2}$Authors are with Universal Robots A/S
        {\tt\small krzi, rype, brbl@universal-robots.com}}%
}

\begin{document}

\maketitle
\copyrightnotice
\thispagestyle{empty}
\pagestyle{empty}

%%%%%%%%%%%%%%%%%%%%%%%%%%%%%%%%%%%%%%%%%%%%%%%%%%%%%%%%%%%%%%%%%%%%%%%%%%%%%%%%
\begin{abstract}
% Deployment of a robotic automation solution is not a simple task. It requires system integration knowledge and experience that the typical robot user lacks. That is one of the main reasons why small and medium enterprises opt-out of automation solutions.
Improving robot deployment is a central step towards speeding up robot-based automation in manufacturing. A main challenge in robot deployment is how to best place the robot within the workcell. To tackle this challenge, we combine two knowledge sources: robotic knowledge of the system and workcell context awareness of the user, and intersect them with an Augmented Reality interface. RobotGraffiti is a unique tool that empowers the user in robot deployment tasks. One simply takes a 3D scan of the workcell with their mobile device, adds contextual data points that otherwise would be difficult to infer from the system, and receives a robot base position that satisfies the automation task. The proposed approach is an alternative to expensive and time-consuming digital twins, with a fast and easy-to-use tool that focuses on selected workcell features needed to run the placement optimization algorithm. The main contributions of this paper are the novel user interface for robot base placement data collection and a study comparing the traditional offline simulation with our proposed method. We showcase the method with a robot base placement solution and obtain up to 16 times reduction in time.
\end{abstract}

\section{INTRODUCTION}
\label{sec:introduction}

%%% Relevant problem to address
The majority of the manufacturing industry consists of Small and Medium Enterprises (SMEs) that are typically characterized by High-Mix Low-Volume production. Due to the variety of manufactured products, these processes are usually not optimized or automated and are thus labor-intensive. Examples of these tasks include manual machine loading and unloading, bin picking, and part inspection. Flexible robotic platforms are designed for dealing with dynamic and rapidly changing manufacturing environments~\cite{article:small_batch}. However, a critical step in the deployment of any robotic manipulator is finding the optimal robot position~\cite{son_convex_2019,feddema_kinematically_1996,kot_method_2021}. As a basic requirement, the robot must have enough reach to complete the task while being able to avoid obstacles in the environment. The impact goes further than this, as the base position can have a significant impact on the overall efficiency of the system. Depending on the optimization goal, a carefully picked base position can help reduce cycle time, increase throughput, and slow down the wear-and-tear of the robot.

\begin{figure}[t]
	\begin{center}
		\includegraphics[width = .99\columnwidth]{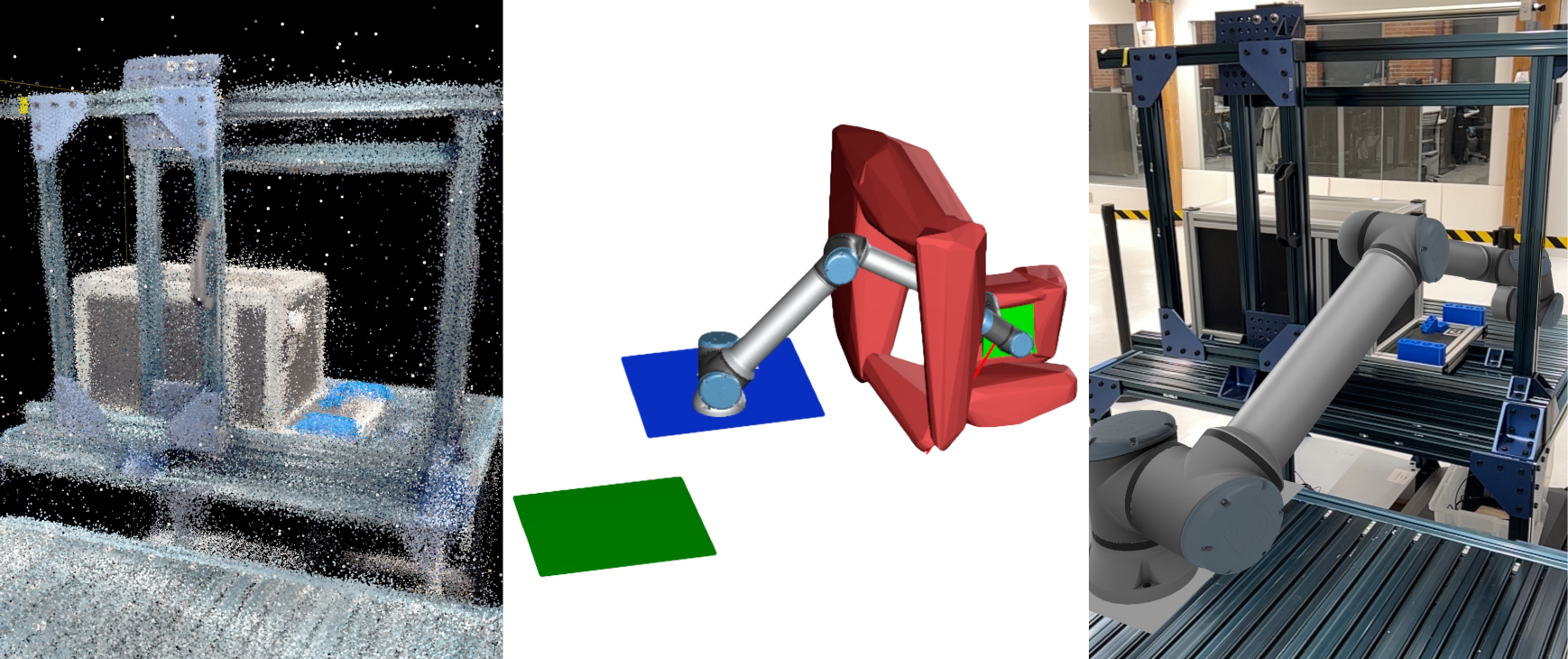}
		\caption{Proposed method. \textit{Left} Scanned point cloud of the workcell, \textit{Center} Collected contextual data used for optimization, \textit{Right} Virtual robot placed in the real workcell.}
		\label{fig:proposed_method}
	\end{center}
\end{figure}

%%% Current steps needed to find the optimal robot position
% \textit{Current steps needed to find the optimal robot position}\\
Finding the optimal robot position requires careful analysis of the production environment, as well as consideration of the specific task that the robot will be performing. However, a person with expert knowledge (such as a system integrator) often makes an educated choice of placement of the robot~\cite{tola_towards_2022}. While this solution might work, it cannot be assumed optimal and limits the flexibility of the system (as it requires assessment from a knowledgeable person every time the system changes). On the other hand, to apply existing methods for optimization of the robot position, a set of parameters needs to be collected and transferred from the real-world workcell into a simulation (or in a more advanced solution - a digital twin). Moreover, the back-and-forth transfer of these parameters accumulates into an error known as the simulation-reality gap: the difference between the real world and the virtual. As this gap grows larger, the utility of the simulation decreases. This problem can be escalated further if one does not have CAD drawings of the real workcell and needs to measure it with a tape measure, increasing the simulation-reality gap. Furthermore, developing and maintaining a digital twin can require significant resources and expertise, which may be a barrier for organizations with limited resources. While digital twins can bring significant benefits (i.e., improved efficiency, reduced downtime and maintenance costs, faster development and testing), they require notable initial investment that can be questionable at best, when the solution constantly needs to be adapted to changed production needs.

The development of Augmented Reality (AR) technology has already made an impact in the field of robotics~\cite{makhataeva_augmented_2020}. AR has also found its use in the deployment of robots~\cite{schuppstuhl_usability_2023}. By overlaying digital information onto the real world, AR can help operators and technicians visualize the robot in its intended environment and identify potential issues before deployment~\cite{bassyouni_augmented_2021}. AR can also be used to provide real-time feedback during deployment, allowing operators to make adjustments and optimize robot performance. One of the commercial solutions, Native Robotics OmniFit~\cite{website:native_robotics} enables users to visualize a robot or a work cell and adds some level of configurability - allowing for removable grippers and additional elements like pallets. However, current approaches to AR have not helped during the real deployment of the robot.

%%%Objective
This paper proposes a new AR-based tool for supporting robot deployment named RobotGraffiti that enables the user to create a rapid model of the workcell using an Augmented Reality interface based on a novel idea for a point cloud paint spraying tool. The goals of this paper are as follows:
\begin{itemize}
    \item remove the need to manually create a simulated version of a workcell to find a satisfactory robot position and hence reduce the deployment time and cost,
    \item provide evidence for the time advantage of using the proposed method over the more traditional method - offline simulation.
\end{itemize}

%%% Contributions
This paper presents two main contributions related to robot deployment. Firstly, a technical solution is proposed for satisfactory robot placement using AR technology in \autoref{sec:method}. Secondly, a comparative study in \autoref{sec:results} evaluates the time and parameter count of the proposed AR-based solution compared to the traditional simulation method, to demonstrate the advantages of using AR in this context. The comparative study is based on expert users as these would be the fastest users of the simulation approach and therefore the best comparison group.

\section{RELATED WORK}
\label{sec:related_work}

\noindent The relevant related work includes the following topics:
\begin{itemize}
    \item virtual deployment and offline robot programming,
    \item point cloud data annotation.
    % \item optimization methods for robot placement.
\end{itemize}

\subsection{Virtual deployment and offline robot programming}
% \todo{handbook of robotics}
Virtual deployment (or commissioning) of the robots is often portrayed as a solution that can reduce downtime, and increase the throughput and efficiency of the robots~\cite{mazumder_towards_2023}. This can be achieved by transferring (or creating) a virtual model of the workcell to be simulated. With an offline programming approach, robot solutions can be tested without damaging the robot and downtime of the operated machines. However, digital twins come with a high initial cost in terms of creating a model. Therefore, they are not typically used for just a single purpose, but often extend the usage from planning into operation. One of the fundamental challenges of digital twins is the simulation-reality gap. To overcome this issue, Tanz et al.~\cite{tanz_automated_2022} suggest using image features of the real-world workcell as additional references to compensate for deviations from the offline-generated robot program. Bousmalis et al.~\cite{bousmalis_using_2017} look at how simulated data can improve learning-based robot grasping systems. The authors find that using pixel-level adaptation in their GAN (Generative Adversarial Network) can improve grasping recognition and reduce the number of real-world images needed. Atorf et al.~\cite{atorf_framework_2017} introduce simulation-based optimization that directly transfers optimization results to the real-world workcell.

%Another issue for virtual deployment is posed when the workcell contains robots from different manufacturers - oftentimes, manufacturers develop proprietary software that supports just their robot models. Andrei et al.~\cite{andrei_perspectives_2021} identify that state-of-the-practice robot simulators allow for multi-manufacturer robot simulation as long as the entire system uses the simulator's manufacturer hardware. Similarly, Reiser et al.~\cite{reiser_real-time_nodate} uses OPC UA and multi-domain modeling language to communicate between different real robots and their virtual twins.

Additionally, some papers focus on displaying digital twin data in Virtual and Augmented Reality. Pérez et al.~\cite{perez_digital_2020} use VR to display virtual workcell with an added level of immersion. Zhu et al.~\cite{zhu_visualisation_2019} present an Augmented Digital Twin that combines real-world workcell enhanced with virtual data from the digital twin. The authors argue that this approach can improve decision-making and higher-level machine control.

\subsection{Point cloud data annotation}
3D point cloud annotation is an inherent field of research connected with machine learning. Many large datasets (KITTI~\cite{geiger_vision_2013}, VMR-Oakland~\cite{munoz_contextual_2009}) exist that are used for training, testing and comparing machine learning algorithms. Since these datasets are used as ground truth models they need to be very accurate, and generation of such datasets is done either manually or semi-manually. Manual annotation of thousands of point clouds is monotonous and time-consuming, so there have been many efforts to improve the annotation efficiency. Yu et al.~\cite{yu_cast_2016} suggest 3 point cloud selection techniques that use tools known from image editing software, such as lasso selection together with heuristics to predict user's intentions. Another approach is used by Monica et al.~\cite{monica_multi-label_2017} that utilizes mouse to point on the point cloud and segmentation strategy (neighborhood graph) to select neighboring points. One of the biggest identified challenges in point cloud annotation is working with 3D dataset on 2D display and input method. Bacim et al.~\cite{bacim_slice-n-swipe_2014} propose "Slice-n-Swipe" with 3DOF (Degree Of Freedom) input method that allows the user to slice unwanted parts of the point cloud with hand gesture tracked with 3D camera. A similar concept - "Go'Then'Tag"~\cite{veit_gothentag_2014} uses a smartphone as a 6DOF input device that acts as a virtual pencil (a raycast with an attached primitive shape) that can mark 3D points on a screen. A lot of efforts have been also put in annotation methods employing Virtual Reality, essentially upgrading input method and display into 3D that allows user unconstrained movement and point selection in the 3D environment. Wilkes et al.~\cite{wilkes_3d_2012} use a smartphone to move objects in a virtual environment. Lin et al.~\cite{lin_immersive_2023} and Wirth et al.~\cite{wirth_pointatme_2019} label points with 3D primitives placed with VR controllers. Franzluebbers et al.~\cite{franzluebbers_virtual_2022} directly select the points with VR controllers, while Lubos et al.~\cite{lubos_touching_2014} use user's hands tracked by 3D camera to de/select points of interest.

This paper draws inspiration from both categories of work and adopts it within the robotic domain. The proposed approach is novel by reducing the complexity of creating a virtual copy of the real environment by enabling users to directly annotate real-world features of the workcell.

\section{METHOD}
\label{sec:method}

\begin{figure*}[h]
    \begin{center}
        \includegraphics[width=\textwidth]{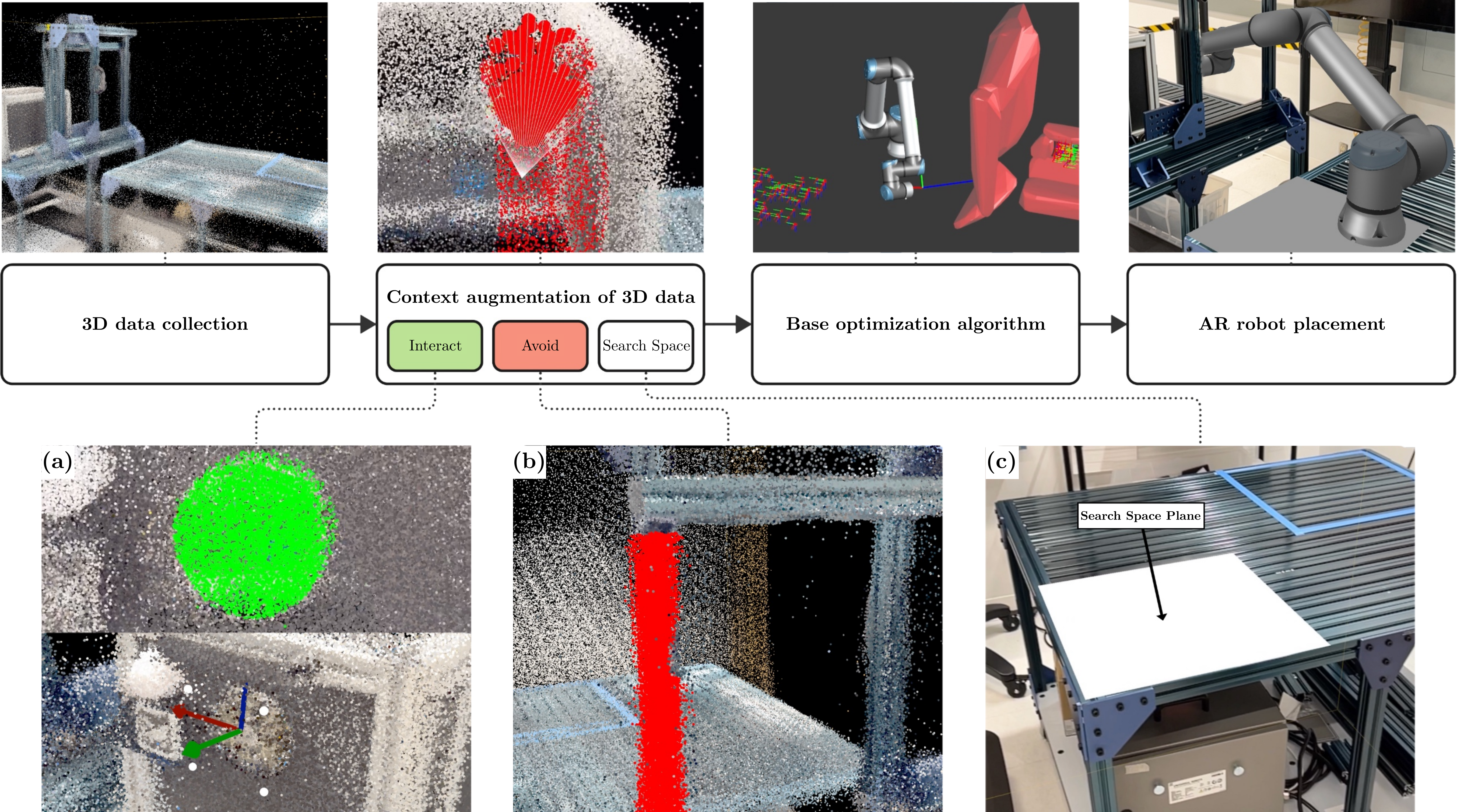}
        \caption{Method workflow with examples of 3 types of context augmentation: \textit{(a)} Top: Sprayed interaction area, Bottom: Calculated corners of the bounding box (marked with white spheres) and orientation (marked with 3D coordinate system model) of the robot task; \textit{(b)} Red points used for collision avoidance in optimization algorithm; \textit{(c)} Search space plane for optimization algorithm.}
        \label{fig:workflow}
    \end{center}
\end{figure*}

The proposed method has the following general workflow for determining a new satisfactory robot placement, as seen in \autoref{fig:workflow}, and listed below:
\begin{enumerate}[label=\emph{\Alph*}.]
    \item 3D data collection - using a device equipped with a LiDAR scanner, we collect a point cloud model of the real-world workcell,
    \item Context augmentation of 3D data - with the RobotGraffiti tool, a user annotates elements in the workcell needed for task execution,
    \item Base optimization algorithm - with the provided input, the method has all the parameters needed to run an optimization algorithm of the choice,
    \item AR robot placement - the system returns the satisfactory robot placement suggestion (if exists) using the AR interface.
\end{enumerate}

\subsection{3D Data Collection}
To make the approach widely applicable we assume a consumer-grade hardware - a mobile device equipped with LiDAR. For a single view, an RGB image and depth map (from LiDAR) are captured. Based on the defined grid size, RGB and depth values are sampled from the image plane. Then, the RGB-D point is projected to 3D world space based on the camera's projection transform. A new pair of RGB image and depth map is captured whenever the camera either changes orientation by $1^{\circ}$ or moves by $0.01 m$. The new points are then added to existent point cloud data. To reduce the volume of data and only capture the area of interest, the user is first asked to set up a bounding box of the workcell.

\subsection{Context Augmentation of 3D Data}
In this step, the user provides the system with context information about the workcell by interacting with previously collected 3D data. We wanted to create an interaction that feels similar to how one would mark surfaces in the real world. To this end, we have developed a tool that marks the points within its' reach, just as paint spraying graffiti, hence the tool name - RobotGraffiti. To spray the points, the user moves the device to the desired points using the AR interface, and then by touching the screen and moving the device, an area can be marked.

RobotGrafiti provides input for rapid workcell modeling, but in this paper is used specifically for the base optimization algorithm, with the following parameters: interaction areas, avoidance areas, and optimization search space. Two of these parameters are provided with two different color sprays:\\
\noindent \green{Green} (\textbf{interact}) - this color is used to mark areas that the robot should interact with. After spraying, a bounding box is calculated with a sphere visualized in each corner - \autoref{fig:workflow}(a). RobotGraffiti does not "spray" the surface but penetrates the point cloud given the dispersed points in the cloud due to sensor uncertainty, so after filtering outliers, it returns the volume which is then converted into a bounding box. Additionally, a 3D coordinate system representation is displayed at the center of the created bounding box that points with a positive Y axis towards the user. The user can set the desired orientation of the robot task (the direction the robot should approach it) by moving the mobile device in the real world with respect to the placement of virtual content. Once satisfied, the user saves the orientation and bounding box points. There can be multiple independent green areas with respective 3D coordinate systems, each representing a different robot task. The information that is given to the optimization algorithm (for every independent area) is as follows: a center point and corners of the bounding box, and a direction vector.\\
\noindent \red{Red} (\textbf{avoid}) - this color is used to mark the areas that the robot should avoid - \autoref{fig:workflow}(b). The sprayed points are reduced to inliers which are then used to generate convex hulls, generated with Qhull algorithm~\cite{barber_quickhull_1996}, that are later used for collision avoidance in the optimization algorithm. A user can add as many red areas as one wants and they can overlap. The red areas are passed as convex hulls with vertices and triangles.

Besides the spraying tool, the user also defines the permissible search space for the robot base. We assume that the user is knowledgeable about the workcell and can provide some general insight into where the robot should be placed. With this approach, the search space can be significantly limited hence reducing the optimization run time. In case the optimization algorithm cannot find a solution, the user can extend or move the search space, or perform the analysis with a robot with a larger reach. In this step, the tool suggests placement of the search space plane on a flat surface (horizontal or vertical), while the user can move, rotate, or change its extent. In the example shown on \autoref{fig:workflow}(c), a plane is placed in the corner of the workbench which is used to align the real and virtual world. The search space plane's center and its extent are passed to the optimization algorithm. Moreover, the center point of the search space plane is used as the origin of the coordinate system for the entire point cloud of the workcell.

\subsection{Base Optimization Algorithm}
The input data provided to the optimization algorithm corresponds to only 3 parameters that were previously determined, discarding the remainder of the point cloud. Simulation software can be chosen freely but must be adapted to the given inputs. Similarly, based on the user's needs, a specific optimization algorithm can be chosen. In this paper (\autoref{sec:results}), we use Advanced Motion Platform~\cite{website:actin}, Universal Robots' proprietary simulation engine with integrated Nelder-Mead optimizer~\cite{neldermead}. If the optimization algorithm fails to return an acceptable result, the user is given the possibility to adjust the search space and rerun the optimization.

% In this paper, we perform base position optimization using a purely kinematic model of the robot and workcell, although future work could integrate dynamics with this approach to more accurately model cycle time and payload capacities.

%%% Simulation and Robot Control
% \noindent\textit{Simulation and Robot Control}\\

% Briefly, this approach attempts to solve:

% \begin{equation}
% \label{eq:ik}
% \mathbf{\dot{q}}=\biggl[\frac{\mathbf{J}}{\mathbf{N_J}^T \mathbf{W}}\biggl]^{-1}\biggl[\frac{\mathbf{V}}{-\alpha \mathbf{N_J}^T{F}}\biggl]
% \end{equation}

% This approach attempts to obtain a joint rate \(\mathbf{\dot{q}}\) given a desired Cartesian tool velocity \(\mathbf{V}\).  The quantity \(\mathbf{N_J}^T\) is a set of vectors spanning the null space of \(\mathbf{J}\), The quantities \(\alpha\), \(\mathbf{W}\) and \(\mathbf{F}\) are parameters that can be defined in a number of ways to influence the solution away from undesirable configurations such as singularities, collisions, and joint limits, provided the system is underdetermined. Here, we use it to avoid both collisions and joint limits. 

\subsection{AR robot placement}
Once the optimization reaches a satisfactory result, the last step of the workflow is to place a virtual robot in the real-world workcell. The base optimization algorithm returns a center point of the robot base in relation to the origin of the coordinate system (the center point of the search space plane), so the virtual robot can be easily placed in the real-world workcell, as seen in \autoref{fig:workflow}.

\section{RESULTS}
\label{sec:results}

To provide the evidence for the contributions listed in the \autoref{sec:introduction}, we conduct an experiment comparing the proposed method with the more traditional approach - offline simulation. The evaluation criteria is the time it takes to obtain optimal robot placement.

%%% Workcell setup
% \noindent\textit{Workcell setup}
\subsection{Workcell setup}
\begin{figure}[h]
	\begin{center}
		\includegraphics[width = .99\columnwidth]{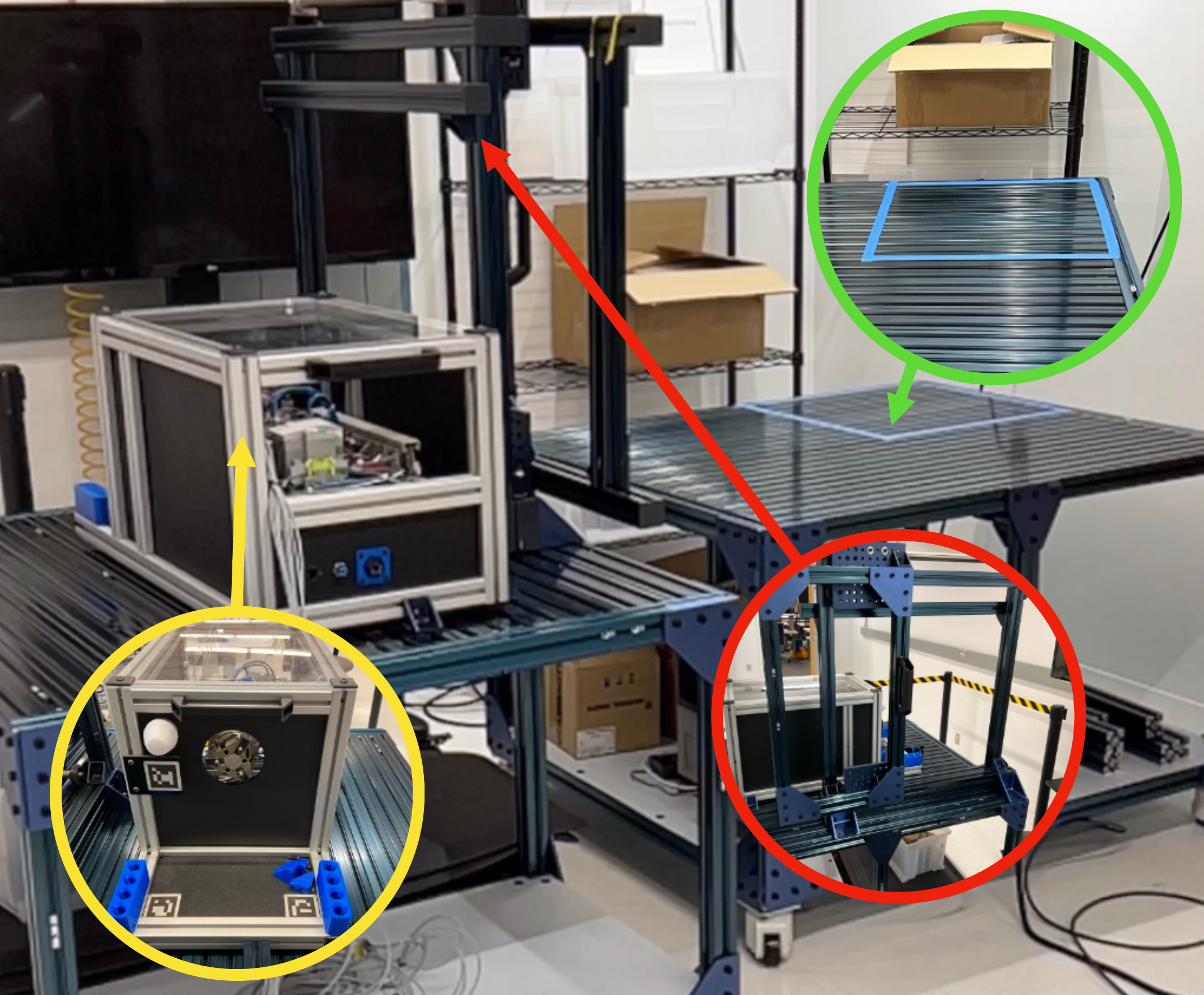}
		\caption{Experiment setup. Elements of the workcell: \textit{green} - tabletop picking area, \textit{yellow} - chuck of the CNC machine, \textit{red} - door of the CNC machine.}
		\label{fig:workcell}
	\end{center}
\end{figure}

The workcell setup for the experiment is a mock-up of a common automation application, machine loading, and can be seen in \autoref{fig:workcell} and includes the following:
\begin{itemize}
    \item a picking area from where the robot is supposed to pick a raw part - marked with blue tape,
    \item loading spot - chuck of the CNC machine where the part needs to be loaded,
    \item avoidance area - door opening of the CNC machine where the robot must navigate through to place the raw part.
\end{itemize}

\subsection{LiDAR sensor}
In this experiment, we use a LiDAR scanner equipped on an iPhone 13 Pro Max. The manufacturer does not provide detailed specifications but we are projecting 2000 points per frame yielding $8.7mm$ resolution (grid size).

%%% Simulation software and robot control
% \noindent\textit{Simulation software and robot control}
\subsection{Simulation software and robot control}
We use the Advanced Motion Platform~\cite{website:actin} (but similar models exist freely available) for both methods: the proposed approach and offline simulation. It allows us to configure the virtual environments, and run the optimization within the same software package. It also provides modeling of full forward and inverse kinematics and the definition of constraints on the motion of the robot. Inverse kinematics and control are modeled using the velocity formulation discussed in~\cite{velcontrol}.

%%% 3D content
% \noindent\textit{3D content}
\subsection{3D content}
\begin{figure}
    \begin{center}
        \includegraphics[width=0.99\columnwidth]{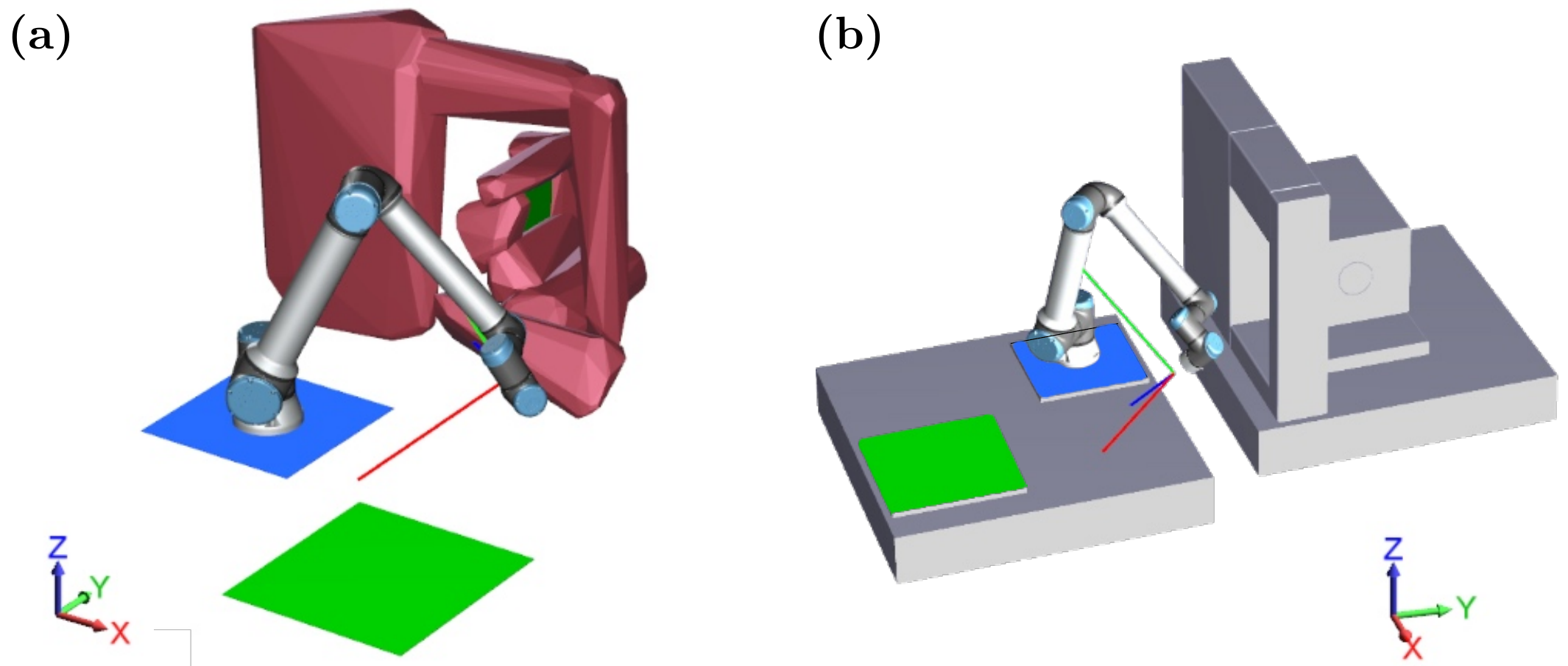}
        \caption{A comparison of the virtual workcell environments generated using \textit{(a)} proposed method, and \textit{(b)} using traditional 3D modelling-based workflow.}
        \label{fig:simcomparison}
    \end{center}
\end{figure}

For the offline simulation, all elements of the workcell (except the robot) are modeled using simple shape primitives to replicate the physical workcell. For the proposed method, the scanned workcell is created using the point cloud generated as described in \autoref{sec:method}. The avoidance zones are imported as convex hulls to simplify the simulation, while interaction zones are used to generate target tool positions. Examples of both simulated workcells are shown in \autoref{fig:simcomparison}.

%%% Workpiece
% \noindent\textit{Workpiece}
\subsection{Workpiece}
We have made the simplifying assumption that the workpiece was symmetric about its vertical axis, and thus could be grasped with any z-rotation of the gripper. This assumption simplifies the optimization by providing an unconstrained DOF that could be used in the control of the manipulator. For this work, the extra DOF is used to avoid collisions with both other elements in the workcell, and the robot itself.

%%% Optimization
% \noindent\textit{Optimization}
\subsection{Optimization}
First, the interaction areas are specified within the simulation, and a set of 100 random target points are selected to be uniformly distributed across each zone (i.e., 100 for the pick zone and 100 for the chuck). Examples of these points are identified with the coordinate frames in \autoref{fig:workflow} (Step 3). We run an optimization algorithm with an objective function defined as:

% The optimization of robot placement was performed over the specified domain by leveraging the simulated control and collision detection systems. First, the interaction areas were specified within the simulation, and a set of 100 random target points were selected to be uniformly distributed across each zone (i.e., 100 for the pick zone and 100 for the chuck). Examples of these points are identified with the coordinate frames in \autoref{fig:simcomparison}(a). The objective function was then defined as:
\begin{equation}
\begin{split}
\begin{aligned}
\max \quad & N(x,y) - 0.1\Bar{D}(x,y) \\
\textrm{s.t.} \quad & \biggl\{ \begin{matrix} -W_x \leq x \leq W_x \\ -W_y \leq y \leq W_y\end{matrix}
\end{aligned}
\end{split}
\end{equation}
\
where \(N\) is the number of points successfully reached, \(\Bar{D}\) is the cumulative total linear distance error from those points that could not be reached and \(W_x\) and \(W_y\) are the extents of the search area. The factor of 0.1 is a scaling factor selected to provide a sufficient gradient toward reaching the most points, while heavily biasing solution towards reaching the maximum number of points. The optimization itself utilizes the NLOpt library~\cite{NLopt} - implementation of the Multi-Level Single-Linkage algorithm~\cite{MLSL}. This is a global optimization algorithm that performs a set of local optimizations based on random starting points and utilizes these results to progress toward a global maximum. These local optimizations are performed using a Nelder-Mead optimizer~\cite{neldermead} due to its simplicity and robustness. It does not require the gradient to be differentiable and it is also unlikely to get stuck in local extrema around singularities.
The result of the optimization is a suggested robot base placement with a percentage of reachable points (taking into consideration collision-free trajectories in between them). In case, the result percentage is below 90\%, the system allows the user to change the search space.

\subsection{Procedure \& Results overview}
\begin{figure}[h]
	\begin{center}
		\includegraphics[width = .99\columnwidth]{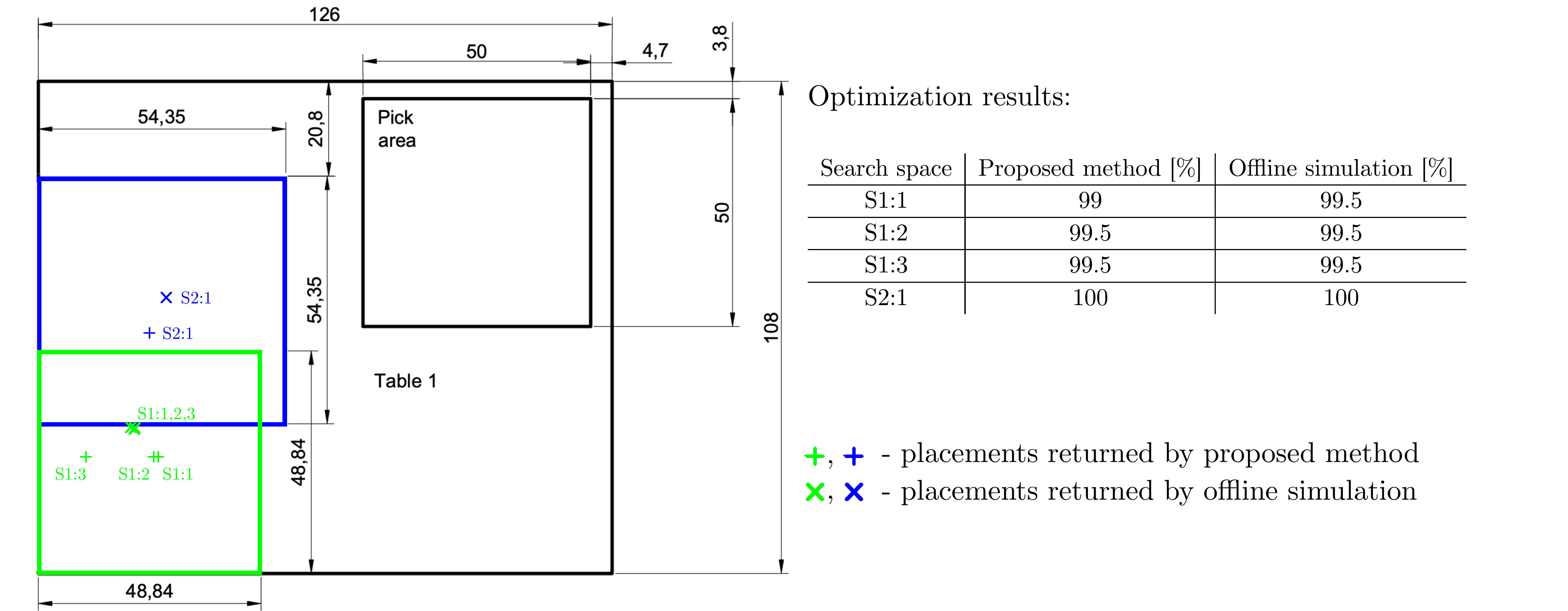}
		\caption{Cutout of top-view 2D CAD drawing of the workcell with 2 search spaces and their placement results from both methods.}
		\label{fig:placement_results}
	\end{center}
\end{figure}

%%% Procedure
We run both methods for the workcell shown in \autoref{fig:workcell} with 2 defined search spaces. We run the methods 4 times (3 in \green{search space 1} to test for repeatability and 1 additional in \blue{search space 2} to test for varying search criteria). The placement results of both methods can be seen in \autoref{fig:placement_results}. Moreover, we enumerate the list of steps and the time it took to complete them for each method, which can be seen in \autoref{tab:steps}. Additionally, we count the amount of individual parameters that are needed to run both of the methods. Those include workcell dimensions, robot tasks, search space area, etc., and are marked with a bold font in the lists. The amount of the parameters and total time needed is presented in the \autoref{tab:comparison}. It is important to stress that the experiment has been conducted with expert users - knowledgeable in 2D CAD design, digital twin, and AR interfaces, hence we can compare the results of both methods.
\newline

%%% Total time
In the conducted experiment, a clear difference in time (\autoref{tab:comparison}) it took for both methods to yield results is visible - over 16 times faster for the proposed method.
%%% Optimization time
The most time-consuming element of the proposed method is running the optimization algorithm, as seen in \autoref{tab:steps}. The reason for the longer processing time is the use of convex hulls for collision avoidance - it is more difficult for the collider to check for collisions with complex mathematical shapes vs. simple primitives used in the offline simulation method. This can be resolved with, i.e., occupancy map that uses cubes which are more efficient for checking collisions, therefore we exclude the processing time from total time for both methods.

%%% Variability
In \autoref{fig:placement_results}, it can be seen that the proposed method has higher variability in the repeatability test. The standard deviation between placement points in \green{search space 1} for the proposed method is $7.3 cm$, while only $0.4 cm$ for the offline simulation.
%%% Optimization results
Moreover, all the tests returned high optimization results (with a maximum of 2 unreachable points), which is above the set threshold, hence the resulting positions yield satisfactory base placement.

\begin{table*}[h]
\caption{List of steps and time needed to complete them for both methods. Time is in seconds.}
\label{tab:steps}
\begin{center}
\begin{tabular}{p{0.39\textwidth}|P{0.06\textwidth}|p{0.39\textwidth}|P{0.06\textwidth}}
\textbf{Proposed method}                                                                          & \textbf{Time [s]} & \textbf{Simulation}                                                                                                                                                                                              & \textbf{Time [s]} \\ \hline
Set bounding box and collect depth data of the workcell                                           & 60             & All the required \textbf{dimensions} are measured with the tape measure and written down on the piece of paper                                                                                  & 900            \\ \hline
Define \textbf{search space area}                                                                          & 15          & 2D CAD drawings are prepared: top-view of the workcell, table, and CNC dimensions & 2700            \\ \hline
Define required \textbf{robot tasks} with RobotGraffiti                          & 45          & Create parameterized \textbf{primitives}                                            & 990            \\ \hline
Define \textbf{obstacle areas} with RobotGraffiti                                                  & 45          & Define required \textbf{robot tasks} - position and orientation                                               & 900            \\ \hline
Place all the collected data into the simulation               & 120            & Define \textbf{search space area}                                                                                                                                                               & 300             \\ \hline
Place \textbf{3D model of the desired robot} with IK solver & 60             & Place \textbf{3D model of the desired robot} with IK solver                                                                                                                & 60             \\ \hline
Run optimization algorithm                                                                        & (450)            & Run optimization algorithm                                                                                                                                                                                       & (60)             \\ \hline
Display virtual robot in the real-workcell                                                        & 5          & Convert result into \textbf{real-world measurement} & 60
\end{tabular}
\end{center}
\end{table*}

\begin{table}[]
\caption{Amount of parameters (measured dimensions, primitives, obstacle areas, etc.) and total time needed to run the simulation and proposed method.}
\label{tab:comparison}
\begin{center}
\begin{tabular}{p{0.11\textwidth}|P{0.145\textwidth}|P{0.115\textwidth}}
                & Amount of parameters & Total time {[}s{]} \\ \hline
Proposed method     & 13                    & 350                     \\ \hline
Simulation          & 76                    & 5910                    
\end{tabular}
\end{center}
\end{table}

\section{DISCUSSION}
\label{sec:discussion}

%%% Time
Even though the proposed method is fast, once set up, simulations can be quickly modified, while the proposed method needs to be reconfigured from the beginning on every change to the workcell. However, in case of a change in the real workcell, the user still must measure the changed dimensions and adjust it accordingly in the simulation. It can be argued that in the time it takes to make those adjustments, the user could scan and augment the data using the proposed approach (it takes $165s$).

%%% Simulation-reality gap
One of the major drawbacks of using offline simulations is the simulation-reality gap that is accumulated with measurement errors fetched into the simulation model, 3D model approximations, and measurement errors when calculating the result. This is also reflected in the amount of parameters needed to prepare a simulation, as seen in \autoref{tab:comparison}. With the proposed approach, these parameters are not needed hence they do not accumulate towards the simulation-reality gap. At the same time, in the proposed approach, the sources of error are sensor measurement errors - LiDAR accuracy affects the collected point cloud, and imprecisely provided context data (sprayed points) from the users - sprayed points might differ from one data collection to another. However, this large variation of input data in the proposed method does not have an impact on the success rate of the solution, as all the placement points returned high reachability result and were tested experimentally to check for the reachability of interaction points.

While discussing the source of errors, human errors in simulation can become more problematic. Besides imprecise 3D model approximation, errors in measurements or omitting important parts/measurements can result in incorrect virtual representation. This can lead to solving a different optimization problem that does not yield the required results. Using our proposed method, the workcell is ensured to be complete, since it is represented by a 3D scan.

%%% Future work
The proposed tool shows significant potential in improving the time it takes to find satisfactory robot base placement but it needs validation with a user study that we intend to conduct in the future. One of the intended limitations of the solution is an area search-space plane to reduce optimization runtime but could be changed into volume search-space. Moreover, the presented method is designed for optimization of the robot placement but a promising direction of future work is to extend it to other robotic (and non-robotic) applications, such as robot programming. RobotGraffiti functionality can be used for other context-driven user data inputs, that can provide the system with more relevant data, otherwise difficult to be implied. Moreover, the tool could be extended to use better fitting primitives than bounding boxes for interaction areas and probabilistic occupancy estimation like OctoMap instead of convex hulls for avoidance areas. This paper evaluates the feasibility of the proposed method, therefore we plan to study it with more complex workcells to evaluate its reliability.

\section{CONCLUSIONS}
\label{sec:conclusions}

This paper proposes a novel method for collecting data to calculate satisfactory robot base placement that ensures the reachability of all the required robot tasks. This method assumes high workcell knowledge of the user - the task to be completed, available space, and available/desired robot. In connection with the awareness of the system of the dimensions of the workcell, a time-consuming and error-prone process of defining robot base placement can be simplified and ensure satisfactory robot placement. This paper introduces a fast (up to 16x faster than the traditional method) and easy-to-use modeling tool which focuses on selected features of the workcell rather than completeness in terms of functionality of the simulation, as opposed to digital twins. It is also simple enough to redo it when the real workcell changes.

\section*{ACKNOWLEDGMENT}
\noindent This paper was produced in collaboration with Universal Robots.
This work is partially funded by Innovation Fund Denmark, as part of Industrial PhD program.

\bibliographystyle{IEEEtran}
\bibliography{IEEEabrv,IEEEexample,mybibliography}

\begin{thebibliography}{10}
\providecommand{\url}[1]{#1}
\csname url@samestyle\endcsname
\providecommand{\newblock}{\relax}
\providecommand{\bibinfo}[2]{#2}
\providecommand{\BIBentrySTDinterwordspacing}{\spaceskip=0pt\relax}
\providecommand{\BIBentryALTinterwordstretchfactor}{4}
\providecommand{\BIBentryALTinterwordspacing}{\spaceskip=\fontdimen2\font plus
\BIBentryALTinterwordstretchfactor\fontdimen3\font minus
  \fontdimen4\font\relax}
\providecommand{\BIBforeignlanguage}[2]{{%
\expandafter\ifx\csname l@#1\endcsname\relax
\typeout{** WARNING: IEEEtran.bst: No hyphenation pattern has been}%
\typeout{** loaded for the language `#1'. Using the pattern for}%
\typeout{** the default language instead.}%
\else
\language=\csname l@#1\endcsname
\fi
#2}}
\providecommand{\BIBdecl}{\relax}
\BIBdecl

\bibitem{article:small_batch}
M.~Löfving, P.~Almström, C.~Jarebrant, B.~Wadman, and M.~Widfeldt,
  ``Evaluation of flexible automation for small batch production,''
  \emph{Procedia Manufacturing}, vol.~25, pp. 177--184, 2018, proceedings of
  the 8th Swedish Production Symposium (SPS 2018).

\bibitem{son_convex_2019}
S.-W. Son and D.-S. Kwon, ``\BIBforeignlanguage{en}{A convex programming
  approach to the base placement of a 6-{DOF} articulated robot with a
  spherical wrist},'' \emph{\BIBforeignlanguage{en}{The International Journal
  of Advanced Manufacturing Technology}}, vol. 102, no. 9-12, pp. 3135--3150,
  Jun. 2019.

\bibitem{feddema_kinematically_1996}
J.~Feddema, ``\BIBforeignlanguage{en}{Kinematically optimal robot placement for
  minimum time coordinated motion},'' in
  \emph{\BIBforeignlanguage{en}{Proceedings of {IEEE} {International}
  {Conference} on {Robotics} and {Automation}}}, vol.~4.\hskip 1em plus 0.5em
  minus 0.4em\relax Minneapolis, MN, USA: IEEE, 1996, pp. 3395--3400.

\bibitem{kot_method_2021}
T.~Kot, Z.~Bobovský, A.~Vysocký, V.~Krys, J.~Šafařík, and R.~Ružarovský,
  ``\BIBforeignlanguage{en}{Method for {Robot} {Manipulator} {Joint} {Wear}
  {Reduction} by {Finding} the {Optimal} {Robot} {Placement} in a {Robotic}
  {Cell}},'' \emph{\BIBforeignlanguage{en}{Applied Sciences}}, vol.~11, no.~12,
  p. 5398, Jun. 2021.

\bibitem{tola_towards_2022}
D.~Tola, E.~Madsen, C.~Gomes, L.~Esterle, C.~Schlette, C.~Hansen, and P.~G.
  Larsen, ``\BIBforeignlanguage{en}{Towards {Easy} {Robot} {System}
  {Integration}: {Challenges} and {Future} {Directions}},'' in
  \emph{\BIBforeignlanguage{en}{2022 {IEEE}/{SICE} {International} {Symposium}
  on {System} {Integration} ({SII})}}.\hskip 1em plus 0.5em minus 0.4em\relax
  Narvik, Norway: IEEE, Jan. 2022, pp. 77--82.

\bibitem{makhataeva_augmented_2020}
Z.~Makhataeva and H.~Varol, ``\BIBforeignlanguage{en}{Augmented {Reality} for
  {Robotics}: {A} {Review}},'' \emph{\BIBforeignlanguage{en}{Robotics}},
  vol.~9, no.~2, p.~21, Apr. 2020.

\bibitem{schuppstuhl_usability_2023}
L.~A. Wulff, M.~Brand, and T.~Schüppstuhl, ``\BIBforeignlanguage{en}{Usability
  of {Augmented} {Reality} {Assisted} {Commissioning} of {Industrial} {Robot}
  {Programs}},'' in \emph{\BIBforeignlanguage{en}{Annals of {Scientific}
  {Society} for {Assembly}, {Handling} and {Industrial} {Robotics} 2022}},
  T.~Schüppstuhl, K.~Tracht, and J.~Fleischer, Eds.\hskip 1em plus 0.5em minus
  0.4em\relax Cham: Springer International Publishing, 2023, pp. 91--102.

\bibitem{bassyouni_augmented_2021}
Z.~Bassyouni and I.~H. Elhajj, ``\BIBforeignlanguage{en}{Augmented {Reality}
  {Meets} {Artificial} {Intelligence} in {Robotics}: {A} {Systematic}
  {Review}},'' \emph{\BIBforeignlanguage{en}{Frontiers in Robotics and AI}},
  vol.~8, p. 724798, Sep. 2021.

\bibitem{website:native_robotics}
``{Native Robotics: Omnifit},'' [Online], Native Robotics, accessed:
  \url{https://native-robotics.com} on 2023-09-13.

\bibitem{mazumder_towards_2023}
A.~Mazumder, M.~Sahed, Z.~Tasneem, P.~Das, F.~Badal, M.~Ali, M.~Ahamed,
  S.~Abhi, S.~Sarker, S.~Das, M.~Hasan, M.~Islam, and M.~Islam,
  ``\BIBforeignlanguage{en}{Towards next generation digital twin in robotics:
  {Trends}, scopes, challenges, and future},''
  \emph{\BIBforeignlanguage{en}{Heliyon}}, vol.~9, no.~2, p. e13359, Feb. 2023.

\bibitem{tanz_automated_2022}
L.~Tanz and R.~Daub, ``\BIBforeignlanguage{en}{Automated {Commissioning} of
  {Offline}-{Generated} {Robot} {Programs}},''
  \emph{\BIBforeignlanguage{en}{Procedia CIRP}}, vol. 107, pp. 810--814, 2022.

\bibitem{bousmalis_using_2017}
K.~Bousmalis, A.~Irpan, P.~Wohlhart, Y.~Bai, M.~Kelcey, M.~Kalakrishnan,
  L.~Downs, J.~Ibarz, P.~Pastor, K.~Konolige, S.~Levine, and V.~Vanhoucke,
  ``\BIBforeignlanguage{en}{Using {Simulation} and {Domain} {Adaptation} to
  {Improve} {Efficiency} of {Deep} {Robotic} {Grasping}},'' Sep. 2017,
  arXiv:1709.07857 [cs].

\bibitem{atorf_framework_2017}
L.~Atorf, C.~Schorn, J.~Rossmann, and C.~Schlette, ``\BIBforeignlanguage{en}{A
  framework for simulation-based optimization demonstrated on reconfigurable
  robot workcells},'' in \emph{\BIBforeignlanguage{en}{2017 {IEEE}
  {International} {Systems} {Engineering} {Symposium} ({ISSE})}}.\hskip 1em
  plus 0.5em minus 0.4em\relax Vienna, Austria: IEEE, Oct. 2017, pp. 1--6.

\bibitem{perez_digital_2020}
L.~Pérez, S.~Rodríguez-Jiménez, N.~Rodríguez, R.~Usamentiaga, and D.~F.
  García, ``\BIBforeignlanguage{en}{Digital {Twin} and {Virtual} {Reality}
  {Based} {Methodology} for {Multi}-{Robot} {Manufacturing} {Cell}
  {Commissioning}},'' \emph{\BIBforeignlanguage{en}{Applied Sciences}},
  vol.~10, no.~10, p. 3633, May 2020.

\bibitem{zhu_visualisation_2019}
Z.~Zhu, C.~Liu, and X.~Xu, ``\BIBforeignlanguage{en}{Visualisation of the
  {Digital} {Twin} data in manufacturing by using {Augmented} {Reality}},''
  \emph{\BIBforeignlanguage{en}{Procedia CIRP}}, vol.~81, pp. 898--903, 2019.

\bibitem{geiger_vision_2013}
\BIBentryALTinterwordspacing
A.~Geiger, P.~Lenz, C.~Stiller, and R.~Urtasun,
  ``\BIBforeignlanguage{en}{Vision meets robotics: {The} {KITTI} dataset},''
  \emph{\BIBforeignlanguage{en}{The International Journal of Robotics
  Research}}, vol.~32, no.~11, pp. 1231--1237, Sep. 2013. [Online]. Available:
  \url{http://journals.sagepub.com/doi/10.1177/0278364913491297}
\BIBentrySTDinterwordspacing

\bibitem{munoz_contextual_2009}
\BIBentryALTinterwordspacing
D.~Munoz, J.~A. Bagnell, N.~Vandapel, and M.~Hebert,
  ``\BIBforeignlanguage{en}{Contextual classification with functional
  {Max}-{Margin} {Markov} {Networks}},'' in \emph{\BIBforeignlanguage{en}{2009
  {IEEE} {Conference} on {Computer} {Vision} and {Pattern}
  {Recognition}}}.\hskip 1em plus 0.5em minus 0.4em\relax Miami, FL: IEEE, Jun.
  2009, pp. 975--982. [Online]. Available:
  \url{https://ieeexplore.ieee.org/document/5206590/}
\BIBentrySTDinterwordspacing

\bibitem{yu_cast_2016}
\BIBentryALTinterwordspacing
L.~Yu, K.~Efstathiou, P.~Isenberg, and T.~Isenberg,
  ``\BIBforeignlanguage{en}{{CAST}: {Effective} and {Efficient} {User}
  {Interaction} for {Context}-{Aware} {Selection} in {3D} {Particle}
  {Clouds}},'' \emph{\BIBforeignlanguage{en}{IEEE Transactions on Visualization
  and Computer Graphics}}, vol.~22, no.~1, pp. 886--895, Jan. 2016. [Online].
  Available: \url{http://ieeexplore.ieee.org/document/7192726/}
\BIBentrySTDinterwordspacing

\bibitem{monica_multi-label_2017}
\BIBentryALTinterwordspacing
R.~Monica, J.~Aleotti, M.~Zillich, and M.~Vincze,
  ``\BIBforeignlanguage{en}{Multi-label {Point} {Cloud} {Annotation} by
  {Selection} of {Sparse} {Control} {Points}},'' in
  \emph{\BIBforeignlanguage{en}{2017 {International} {Conference} on {3D}
  {Vision} ({3DV})}}.\hskip 1em plus 0.5em minus 0.4em\relax Qingdao: IEEE,
  Oct. 2017, pp. 301--308. [Online]. Available:
  \url{https://ieeexplore.ieee.org/document/8374583/}
\BIBentrySTDinterwordspacing

\bibitem{bacim_slice-n-swipe_2014}
\BIBentryALTinterwordspacing
F.~Bacim, M.~Nabiyouni, and D.~A. Bowman,
  ``\BIBforeignlanguage{en}{Slice-n-{Swipe}: {A} free-hand gesture user
  interface for {3D} point cloud annotation},'' in
  \emph{\BIBforeignlanguage{en}{2014 {IEEE} {Symposium} on {3D} {User}
  {Interfaces} ({3DUI})}}.\hskip 1em plus 0.5em minus 0.4em\relax MN, USA:
  IEEE, Mar. 2014, pp. 185--186. [Online]. Available:
  \url{http://ieeexplore.ieee.org/document/6798882/}
\BIBentrySTDinterwordspacing

\bibitem{veit_gothentag_2014}
\BIBentryALTinterwordspacing
M.~Veit and A.~Capobianco, ``\BIBforeignlanguage{en}{Go'{Then}'{Tag}: {A} 3-{D}
  point cloud annotation technique},'' in \emph{\BIBforeignlanguage{en}{2014
  {IEEE} {Symposium} on {3D} {User} {Interfaces} ({3DUI})}}.\hskip 1em plus
  0.5em minus 0.4em\relax Minneapolis, MN: IEEE, Mar. 2014, pp. 193--194.
  [Online]. Available: \url{http://ieeexplore.ieee.org/document/6798886/}
\BIBentrySTDinterwordspacing

\bibitem{wilkes_3d_2012}
\BIBentryALTinterwordspacing
C.~B. Wilkes, D.~Tilden, and D.~A. Bowman, ``\BIBforeignlanguage{en}{{3D}
  {User} {Interfaces} {Using} {Tracked} {Multi}-touch {Mobile} {Devices}},''
  \emph{\BIBforeignlanguage{en}{Joint Virtual Reality Conference of ICAT - EGVE
  - EuroVR}}, p. 8 pages, 2012, artwork Size: 8 pages ISBN: 9783905674408
  Publisher: The Eurographics Association. [Online]. Available:
  \url{http://diglib.eg.org/handle/10.2312/EGVE.JVRC12.065-072}
\BIBentrySTDinterwordspacing

\bibitem{lin_immersive_2023}
\BIBentryALTinterwordspacing
T.~Lin, Z.~Yu, N.~Volkens, M.~McGinity, and S.~Gumhold,
  ``\BIBforeignlanguage{en}{An {Immersive} {Labeling} {Method} for {Large}
  {Point} {Clouds}},'' in \emph{\BIBforeignlanguage{en}{2023 {IEEE}
  {Conference} on {Virtual} {Reality} and {3D} {User} {Interfaces} {Abstracts}
  and {Workshops} ({VRW})}}.\hskip 1em plus 0.5em minus 0.4em\relax Shanghai,
  China: IEEE, Mar. 2023, pp. 829--830. [Online]. Available:
  \url{https://ieeexplore.ieee.org/document/10108505/}
\BIBentrySTDinterwordspacing

\bibitem{wirth_pointatme_2019}
\BIBentryALTinterwordspacing
F.~Wirth, J.~Quehl, J.~Ota, and C.~Stiller,
  ``\BIBforeignlanguage{en}{{PointAtMe}: {Efficient} {3D} {Point} {Cloud}
  {Labeling} in {Virtual} {Reality}},'' in \emph{\BIBforeignlanguage{en}{2019
  {IEEE} {Intelligent} {Vehicles} {Symposium} ({IV})}}.\hskip 1em plus 0.5em
  minus 0.4em\relax Paris, France: IEEE, Jun. 2019, pp. 1693--1698. [Online].
  Available: \url{https://ieeexplore.ieee.org/document/8814115/}
\BIBentrySTDinterwordspacing

\bibitem{franzluebbers_virtual_2022}
\BIBentryALTinterwordspacing
A.~Franzluebbers, C.~Li, A.~Paterson, and K.~Johnsen,
  ``\BIBforeignlanguage{en}{Virtual {Reality} {Point} {Cloud} {Annotation}},''
  in \emph{\BIBforeignlanguage{en}{Proceedings of the 2022 {ACM} {Symposium} on
  {Spatial} {User} {Interaction}}}.\hskip 1em plus 0.5em minus 0.4em\relax
  Online CA USA: ACM, Dec. 2022, pp. 1--11. [Online]. Available:
  \url{https://dl.acm.org/doi/10.1145/3565970.3567696}
\BIBentrySTDinterwordspacing

\bibitem{lubos_touching_2014}
\BIBentryALTinterwordspacing
P.~Lubos, R.~Beimler, M.~Lammers, and F.~Steinicke,
  ``\BIBforeignlanguage{en}{Touching the {Cloud}: {Bimanual} annotation of
  immersive point clouds},'' in \emph{\BIBforeignlanguage{en}{2014 {IEEE}
  {Symposium} on {3D} {User} {Interfaces} ({3DUI})}}.\hskip 1em plus 0.5em
  minus 0.4em\relax MN, USA: IEEE, Mar. 2014, pp. 191--192. [Online].
  Available: \url{http://ieeexplore.ieee.org/document/6798885/}
\BIBentrySTDinterwordspacing

\bibitem{barber_quickhull_1996}
\BIBentryALTinterwordspacing
C.~B. Barber, D.~P. Dobkin, and H.~Huhdanpaa, ``\BIBforeignlanguage{en}{The
  quickhull algorithm for convex hulls},'' \emph{\BIBforeignlanguage{en}{ACM
  Transactions on Mathematical Software}}, vol.~22, no.~4, pp. 469--483, Dec.
  1996. [Online]. Available: \url{https://dl.acm.org/doi/10.1145/235815.235821}
\BIBentrySTDinterwordspacing

\bibitem{website:actin}
``{Universal Robots Actin UR},'' [Online], Universal Robots, accessed:
  \url{https://www.universal-robots.com/fi/plus/products/energid/actin-ur/} on
  2023-09-13.

\bibitem{neldermead}
J.~A. Nelder and R.~Mead, ``{A Simplex Method for Function Minimization},''
  \emph{The Computer Journal}, vol.~7, no.~4, pp. 308--313, 01 1965.

\bibitem{velcontrol}
J.~English and A.~Maciejewski, ``\BIBforeignlanguage{en}{On the implementation
  of velocity control for kinematically redundant manipulators},''
  \emph{\BIBforeignlanguage{en}{IEEE Transactions on Systems, Man and
  Cybernetics}}, vol.~40, no.~3, pp. 233--237, May 2000.

\bibitem{NLopt}
S.~G. Johnson, ``The {NLopt} nonlinear-optimization package,''
  \url{https://github.com/stevengj/nlopt}, 2007.

\bibitem{MLSL}
A.~H. G.~R. Kan and G.~T. Timmer, ``Stochastic global optimization methods part
  {II}: Multi level methods,'' \emph{Mathematical Programming}, vol.~39, pp.
  57--78, 1987.

\end{thebibliography}

%\bibliographystyle{ieeetr}

%\bibliography{mybibliography.bib}

\end{document}